\newif\ifarxiv
\newif\ifralfinal
\ralfinalfalse \arxivtrue

\ifarxiv
\documentclass[letterpaper, 10 pt, conference]{ieeeconf}  
\fi
\ifralfinal
\documentclass[letterpaper, 10 pt, journal, twoside]{IEEEtran}
\fi

\IEEEoverridecommandlockouts                              

\usepackage{graphics} 
\usepackage{epsfig} 
\usepackage{mathptmx} 
\usepackage{times} 
\usepackage{amsmath} 
\usepackage{amssymb}  
\usepackage{booktabs}
\usepackage{amsmath}
\usepackage{hyperref}
\DeclareMathOperator*{\argmax}{arg\,max}
\ifarxiv
\usepackage{fancyhdr}
\fi

\begin{document}

\title{
\ifarxiv\LARGE \bf\fi
A Hierarchical Dual Model of Environment- and Place-Specific Utility for Visual Place Recognition
}

\author{Nikhil Varma Keetha$^{1}$, Michael Milford$^{2}$ and Sourav Garg$^{2}$
\thanks{$^{1}$The author is with the Indian Institute of Technology (ISM) Dhanbad, India.
        }%
\thanks{$^{2}$The authors are with the School of Electrical Engineering and Robotics, QUT, Brisbane, Australia. The authors acknowledge continued support from the Queensland University of Technology (QUT) through the Centre for Robotics.
        }%
}



\maketitle
\ifarxiv
\thispagestyle{fancy}
\pagestyle{plain}
\fi

\begin{abstract}
Visual Place Recognition (VPR) approaches have typically attempted to match places by identifying visual cues, image regions or landmarks that have high ``utility'' in identifying a specific place. But this concept of utility is not singular - rather it can take a range of forms. In this paper, we present a novel approach to deduce two key types of utility for VPR: the utility of visual cues `specific' to an environment, and to a particular place. We employ contrastive learning principles to estimate both the environment- and place-specific utility of Vector of Locally Aggregated Descriptors (VLAD) clusters in an unsupervised manner, which is then used to guide local feature matching through keypoint selection. By combining these two utility measures, our approach achieves state-of-the-art performance on three challenging benchmark datasets, while simultaneously reducing the required storage and compute time. We provide further analysis demonstrating that unsupervised cluster selection results in semantically meaningful results, that finer grained categorization often has higher utility for VPR than high level semantic categorization (e.g. building, road), and characterise how these two utility measures vary across different places and environments. Source code is made publicly available at \url{https://github.com/Nik-V9/HEAPUtil}.
\end{abstract}


\section{Introduction}
\ifralfinal
\else
Mobile
\fi robot localization can be challenging due to extreme variations in scene appearance and camera viewpoint, which affect key robot capabilities including semantic scene understanding and Visual Place Recognition (VPR). Several solutions have been proposed in the literature to improve VPR, including contrastive representation learning~\cite{arandjelovic2016netvlad,revaud2019learning}, domain translation~\cite{latif2018addressing}, sequential matching~\cite{garg2021seqnet,garg2020delta}, semantic saliency~\cite{garg2018lost,naseer2017semantics} and hierarchical matching~\cite{cummins2011appearance,sarlin2019coarse}. Many of these methods tend to learn salient visual cues that can improve VPR, but which are typically non-interpretable~\cite{arandjelovic2016netvlad,detone2018superpoint}, or in the case of those based on explicit semantics, require human input for interpretation and performance enhancement~\cite{garg2018lost,gawel2018x,naseer2017semantics}.

\begin{figure}
\centering
\begin{tabular}{cccc}
\includegraphics[width=0.4\textwidth]{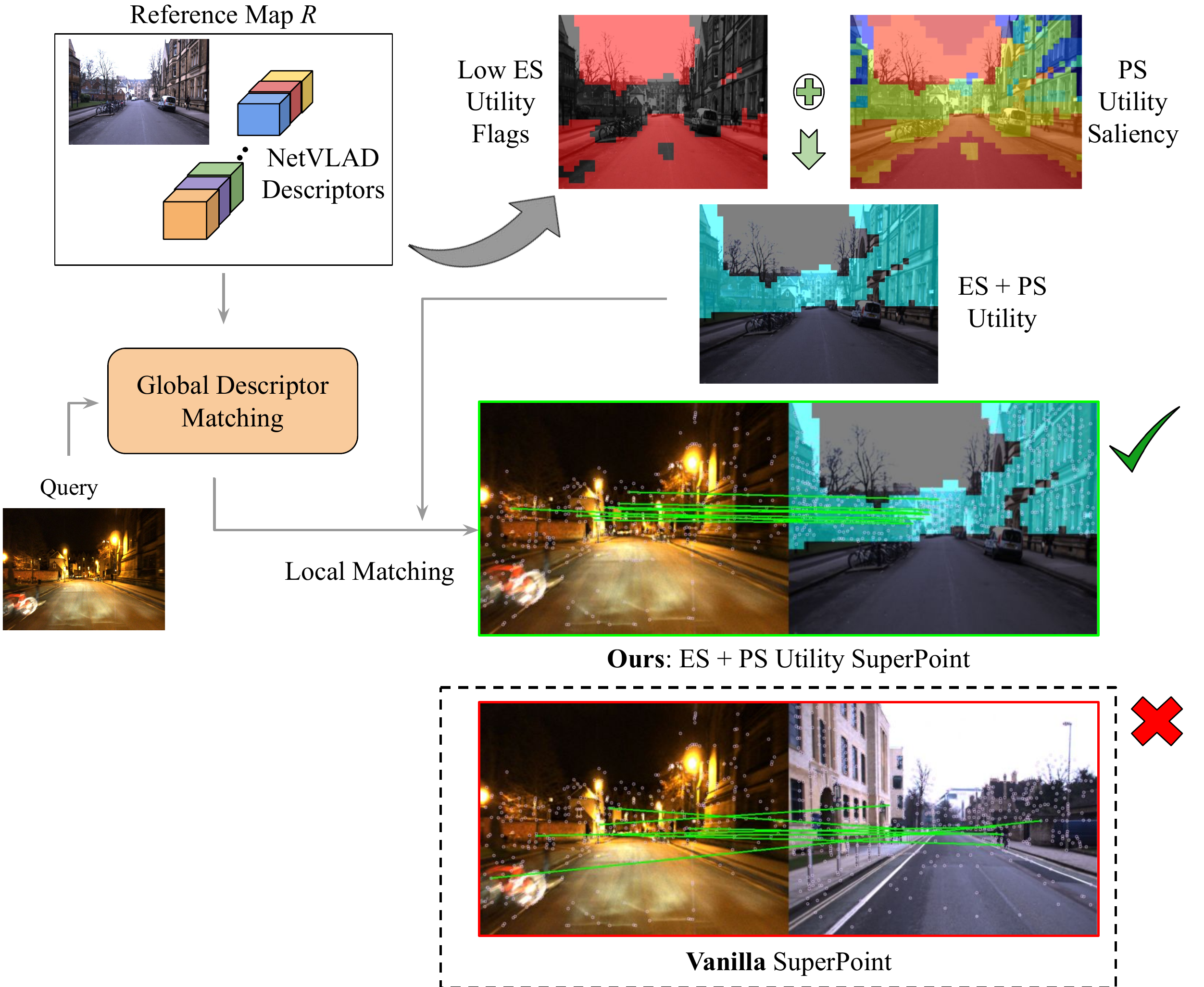} \\
\end{tabular}
\caption{We propose Environment-Specific (ES) and Place-Specific (PS) utility estimation methods that determine the relevance of unique visual cues in a reference map. Our combined ES and PS utility, estimated from the global NetVLAD descriptors, guides SuperPoint's keypoint selection (cyan mask) to obtain correct feature correspondences, where vanilla SuperPoint fails due to matches found on pedestrians and vehicles.}
\label{fig:Overview}
\end{figure}

In this work, we propose novel visual utility estimation techniques that not only lead to state-of-the-art VPR performance, but also offer semantic interpretability. For example, they enable insights into the role of coarse (buildings) vs fine-grained semantics (parts of buildings), as well as the varying relevance of a particular semantic class (road) in different environmental contexts (city vs rail traverse). We present a novel hierarchical VPR pipeline that uses global descriptors to guide local feature matching in a more unified manner. Moving beyond existing hierarchical VPR methods which only pass candidate hypotheses from the coarse to fine stage, we also estimate the visual utility of different elements in the scene through a VLAD-based global descriptor (NetVLAD~\cite{arandjelovic2016netvlad}). In particular, we use the cluster-level VLAD representations to estimate a cluster's utility in \textit{environment-specific} and \textit{place-specific} manner for a given reference map. Here, environment-specific refers to estimating utility at a global level applicable to all the places, whereas place-specific refers to estimating utility at a local level applicable to that particular place only. Without requiring any special iterative training, the proposed method is tested on different environments, where we show that the combination of global environment-specific and local place-specific utility leads to informed local feature matching and reduces storage and compute requirements (see Fig~\ref{fig:Overview}).

We make the following specific contributions:

\begin{itemize}
    \item an unsupervised method for estimating the global environment-specific (ES) and local place-specific (PS) utility of visual elements represented as VLAD clusters;
    \item a \textit{more} unified hierarchical global-to-local VPR pipeline where utility estimated from global descriptors guides local feature matching;
    \item a combined ES and PS utility-based method which achieves state-of-the-art VPR performance while offering reduced storage and compute time properties; and
    \item a `bridge' between human semantics and automated segmentation-based understanding of visual relevance for VPR, achieved through several visualizations and qualitative insights.
\end{itemize}

\ifarxiv
\setlength{\topmargin}{-24pt}
\setlength{\headheight}{0pt}
\fi
\section{Related Work}
\subsection{Global and Local Descriptors for VPR}
\label{sec:LitDescriptors}

VPR is commonly posed as an image retrieval problem, where an image is described by a global descriptor or a set of local descriptors and keypoints to match with other images. Recent surveys~\cite{masone2021survey, garg2021where, lowry2015visual} have reviewed the many representations used to describe images for VPR, ranging from hand-crafted features such as SIFT~\cite{lowe2004distinctive} to learned global descriptors such as NetVLAD~\cite{arandjelovic2016netvlad}, AP-GeM~\cite{revaud2019learning}, and DeLG~\cite{cao2020unifying}; local descriptors such as SuperPoint~\cite{detone2018superpoint} and DeLF~\cite{cao2020unifying}; and local matchers such as SuperGlue~\cite{sarlin2020superglue}.

Furthermore, hierarchical approaches have been used in several VPR, and SLAM pipelines where global descriptors are used to retrieve top candidates and local feature matching is used to obtain the best match amongst the top candidates~\cite{cummins2011appearance,garg2018lost,mur2015orb,engel2014lsd}. One such approach, HF-Net~\cite{sarlin2019coarse} proposed a `monolithic' CNN to simultaneously learn global NetVLAD descriptors and local SuperPoint features for 6-DoF localization. However, in such hierarchical approaches, the global descriptors are not used to guide the local feature matching, and are limited to providing top matching candidates. In this context, our proposed hierarchical method uses unsupervised utility estimated from global descriptors to guide local feature matching, and is applicable to any existing hierarchical VPR method.

\subsection{Visual Feature Selection}
\label{sec:LitSemantics}
The VPR problem has been posed in many ways ranging from a classification task~\cite{chen2017deep,cao2020unifying} to contrastive learning~\cite{arandjelovic2016netvlad,revaud2019learning,radenovic2018fine}. Most methods share the core intuition of attempting to represent images of the same place similarly. However, environments are filled with distractors, and so much work has attempted to automatically learn and identify the areas of an image with the most utility for the VPR task.

\paragraph{Semantics Based} As surveyed recently~\cite{garg2020semantics}, visual semantics is an emerging area of research in the field of robotics with huge potential for VPR and localization. A number of methods have demonstrated the use of semantic information or distinctive and informative visual elements for improving VPR~\cite{schreiber2013laneloc, atanasov2016localization, stone2016skyline, weng2018semantic, naseer2017semantics,garg2018lost,gawel2018x}. However, these methods rely largely on human-based semantic categories, where relevant information is retained based on human intuition of the semantic classes, for example, buildings~\cite{naseer2017semantics}, roads~\cite{garg2018lost,gawel2018x}, lanes~\cite{schreiber2013laneloc} and the skyline~\cite{stone2016skyline}. Such approaches also tend to require segmentation masks or supervision involving semantic labels to endow the system with higher-level semantic knowledge~\cite{naseer2017semantics, gawel2018x, garg2018lost}. More recently, \cite{larsson2019fine} explored using more fine-grained semantic categories beyond those derived from humans, showing promising potential.

\paragraph{Region and Attention Based}
Past methods have adopted region-based approaches, including grid-based region selection, region proposal networks, Hashing based landmark detection, and Convolutional Neural Network (CNN) activations based region extraction~\cite{8202131,khaliq2019holistic,8421024,sunderhauf2015place}. Other approaches involve using attention to weigh image features based on a relevance criterion~\cite{gordo2016deep}.  However, even though semantics and saliency are strongly intertwined, there are no methods that leverage learned descriptors' inherent semantic properties for this problem. In this context, our proposed framework estimates the utility of NetVLAD~\cite{arandjelovic2016netvlad} clusters, leveraging their intrinsic semantic properties.

\paragraph{Place-Specific Feature Selection}
\label{sec:LitDiscriminative} 
Place-specific learning has been explored previously~\cite{carl2012makes, gronat2013learning, mcmanus2014scene} but these methods require significant training. \cite{knopp2010avoiding} proposed an image-specific and spatially-localized detection of confusing features using local Term frequency-Inverse document frequency (tf-idf) weighting, however, it does not consider a simultaneous global environment-level utility as proposed in this work. Furthermore, we show that our method can be employed under challenging appearance conditions and is not limited to city-like environments.

\begin{figure*}
\centering
\begin{tabular}{cccc}
\includegraphics[trim={0.2cm 0cm 0.1cm 0cm},clip,width=0.75\textwidth]{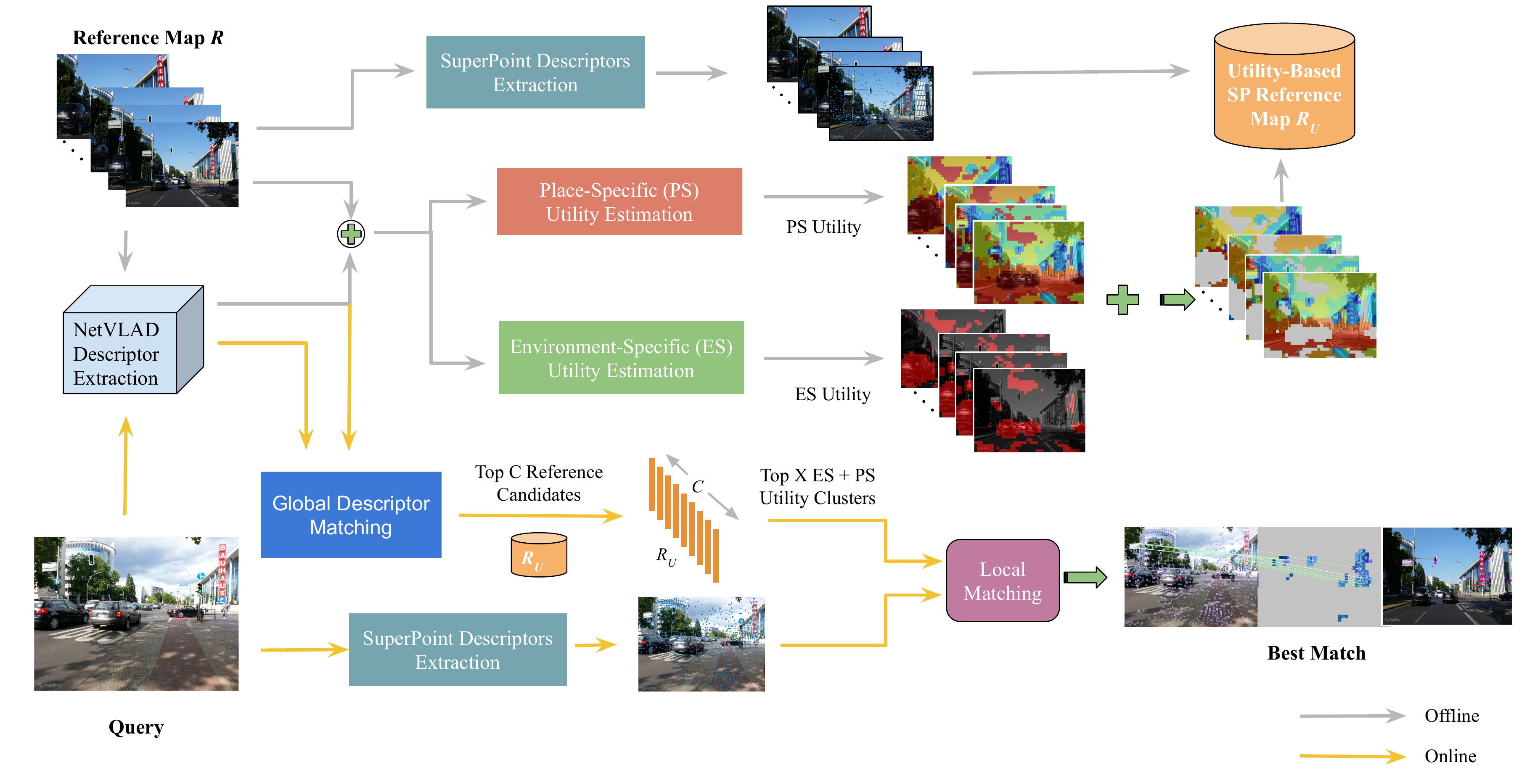} \\
\end{tabular}
\caption{Schematic of our proposed approach. In the offline stage, global and local descriptors are extracted from the reference database images, and environment-specific (ES) and place-specific (PS) utility is estimated to further filter the local keypoints. During the online localization stage, for a given query, the top C matching candidates are retrieved from the reference database using global descriptor matching. The final place match is then obtained through local feature matching of query image features with the high utility features of the candidates.}
\label{fig:Approach}
\end{figure*}

\section{Proposed Approach}
In this section, we first present the proposed unsupervised techniques to estimate environment- and place-specific feature utility for VPR. We then describe our unified approach to hierarchical VPR, where the coarse global descriptor matching stage guides the fine local feature matching stage via keypoint filtering based on utility estimates (see Fig~\ref{fig:Approach}).

\subsection{Feature Utility Estimation}

VLAD based place representations~\cite{jegou2010aggregating,torii201524,arandjelovic2016netvlad,garg2018lost} have been demonstrated to achieve high performance VPR, in particular the recent deep learning based adaptations such as NetVLAD~\cite{arandjelovic2016netvlad}. Our approach here is motivated by the observation that the cluster assignment of NetVLAD descriptors has inherent semantic properties, whose level of detail varies with the number of clusters. For example, a NetVLAD descriptor with $16$ clusters\footnote{$16$ was chosen as being the closest to typical number of semantic classes for road-based datasets such as Cityscapes~\cite{cordts2016cityscapes}. Please refer to Section~\ref{sec:vocabSize} for an ablation study on vocabulary size.} results in cluster assignment that is analogous to human-based semantics while also comprising fine-grained segmentation of typical broad semantic classes like buildings and roads. These observations lead to the hypothesis we pursue here: instead of comparing the full concatenated VLAD descriptor, if we compare the aggregated residuals at cluster level, the distribution of cluster-wise distances within the reference map can be used to estimate that particular cluster's utility for VPR. The second component of this hypothesis is that clusters with lower cluster-wise distances tend to cause high perceptual aliasing. We formulate this procedure in accordance with the well established max-margin based contrastive learning regime, as described in the following subsections.

\subsubsection{Maximizing cluster-wise margins}
Contrastive learning has been demonstrated to achieve state-of-the-art performance for representing places~\cite{arandjelovic2016netvlad,revaud2019learning,radenovic2018fine}. With the use of triplets, that is, an anchor ($X_a$), positive ($X_p$) and a negative ($X_n$), descriptors are typically learnt such that the margin, $\alpha$, between the anchor-negative distance and the anchor-positive distance is maximized. 
\begin{equation}
    \alpha = \lVert X_a-X_n\rVert_2 - \lVert X_a-X_p\rVert_2
\end{equation}

In this work, we maximize this margin in a non-iterative manner since only the reference map traverse is used, unlike the typical use of multiple place views~\cite{arandjelovic2016netvlad, revaud2019learning}. We further adapt the computation of this margin with two key points: $a)$ the triplets are represented with cluster-wise aggregated residuals instead of considering a full concatenated VLAD vector and $b)$ the anchor-positive distance in our case approaches zero, since we only consider a single traverse where nearby images are similar. The margin is thus computed for each cluster independently using only the anchor-negative distances, and is referred to as \textit{utility} from here. Thus, the higher the utility of a cluster, the lower the perceptual aliasing it causes. This process can be used at both the environment-level (that is, globally across the full reference map) and place-level (that is, specific to individual local places), respectively referred to as environment-specific and place-specific, as discussed further here.

\subsubsection{Place-Specific Utility}
\label{sec:PS}
High-utility clusters tend to indicate salient areas of a particular place. Based on this observation, we formulate a per cluster place-specific utility estimation which, when sorted, gives us a relative saliency ranking for each cluster at the particular place in the reference map.

For a given Image $I$ (considered as a unique place) from a particular geographical location in the Reference Map $R$, a positive localization radius $P$ exists where places (images) within this radius are considered positives. Also, a non-negative localization radius $2P$ is considered beyond which all places (images) are considered negatives. Let us suppose that there are $Z_a$ negatives in $R$ for anchor Image $a$ such that $n_1, n_2,\cdots, n_z$ represent each negative. Then the place-specific utility of cluster $K$, given that it exists, at that particular place with anchor image $a$ in $R$ can be formulated as:

\begin{equation}
\centering
    (_{}^{PS}U_K)_a = \frac{1}{Z_a}\sum_{n=n_1}^{n_z} \lVert (V_K)_a - (V_K)_n \rVert_2
\end{equation}
where $V_K$ represents the sum of residuals for the cluster K.

\subsubsection{Environment-Specific Utility}
\label{sec:ES}

At an environment-level, clusters with a low variance in residual values across the reference map contain objects with high perceptual aliasing for that particular environment. We hypothesize that such clusters vary by property of the specific environment and such global utility can guide the place-specific utility to avoid non-relevant clusters for that environment while also preventing transient errors.

Considering all $N$ places\footnote{Unlike SLAM, for global re-localization the map size is known a priori.} in the reference map $R$, the Environment-Specific utility, $_{}^{ES}U_K$, of a cluster $K$, is formulated as:

\begin{equation}
\centering
    _{}^{ES}U_K = \frac{1}{N(N-1)}\sum_{a=1}^{N}\sum_{n=n_1}^{n_z} \lVert (V_K)_a - (V_K)_n \rVert_2
\end{equation}
where $V_K$ represents the cluster-level representation of a place, that is, the sum of residuals, for cluster $K$. Once the utility values for all the $K$ clusters are determined, k-means segregation\footnote{The use of terms `segregation' and `bins' instead of clustering and clusters for k-means is intentional to avoid confusing it with VLAD clusters.} with k=$2$ is used to divide the clusters into two bins. VLAD clusters falling in the bin with high utility are regarded as environment-specific high utility clusters, while others are discarded as dustbin clusters.

\subsection{Unified Hierarchical Visual Place Recognition}
\label{sec:HierApp}

When local feature matching pipelines compare local descriptors, they can struggle to discard distractors within an image or areas of the image with high perceptual aliasing. We propose to increase their robustness to aliasing by filtering the local descriptors and keypoints based on our place-specific and environment-specific utility estimated from global descriptors, with an additional benefit being the reduced storage and compute requirements. Distinct from existing work, our hierarchical VPR better unifies the global and local feature stages as the keypoint utility is \textit{directly} estimated through the utility of VLAD clusters of the global descriptors.

\subsubsection{Global Descriptor Matching}
\label{sec:GDM}
We use NetVLAD representation as our global descriptor. A query image descriptor is matched with the reference global descriptors using Euclidean distance to retrieve top $C$ matching candidates.

In order to obtain a cluster-level segmentation mask, we replace NetVLAD's differentiable soft cluster assignment with its original counterpart of hard cluster assignment, leading to a mask of size $40\times30$ corresponding to the spatial dimensions of NetVLAD's last convolutional layer tensor. This cluster assignment mask is then rescaled to the original image size $640\times480$. These cluster assignments along with the top $C$ candidate matches are then passed on to the local feature matching stage.

\subsubsection{Local Feature Matching}
\label{sec:LFM}
For local feature matching, we use SuperPoint~\cite{detone2018superpoint} (SP) descriptors/keypoints. Given the reference images database $R$, cluster assignment mask from the global descriptors and keypoint spatial locations, we employ the place-specific and environment-specific utility for SP descriptors/keypoints filtering in the following manner:

a) \textit{Environment-Specific keypoint filtering}: Based on the unsupervised environment specific utility estimation, we select the SP keypoints corresponding to environment-specific high utility clusters.

b) \textit{Place-Specific keypoint filtering}: Once the place-specific utility of each cluster for an Image $I$ in the Reference map $R$ is estimated, the utility values of all the clusters are sorted to obtain a relative cluster saliency ranking for that specific Image $I$. Based on this place-specific cluster saliency ranking, we use the \textit{Top X Clusters} to select SP keypoints.

c) \textit{Combined keypoint filtering}: We propose a combination of the environment-specific and place-specific utility approaches. Initially, the SP keypoints of an image are subsampled using the Top X clusters formulation, and then a further filtering is performed to obtain keypoints belonging only to environment-specific high utility clusters.

Once the filtered SP keypoints for all the images in the reference map are obtained, we consider two state-of-the-art feature matching pipelines to match the SP descriptors of the query with the filtered SP descriptors of the top $C$ reference candidates obtained through global descriptor matching:

i) Based on SuperPoint's matching pipeline~\cite{detone2018superpoint}, the descriptors of query and reference image are first matched using absolute Euclidean distance-based Nearest Neighbor (NN) search with mutual NN cross-check, which is followed by geometric verification using RANSAC based homography with a pixel threshold of $3$.

ii) Based on SuperGlue's matching pipeline~\cite{sarlin2020superglue}, a graph neural network takes as input SuperPoint keypoints and descriptors to produce inliers between an image pair.

For a matched image pair, inliers are used to compute the match score:

\begin{equation}
\centering
    c_{final} = \argmax_{c\in C}\frac{p_I}{p_Q+p_c}
\end{equation}

where $p_I$, $p_Q$ and $p_c$ are the number of inliers, the number of SP keypoints in the query, and the number of filtered SP keypoints in the candidate reference image, respectively. Amongst the top $C$ candidates, the candidate with the highest match score is selected as the final match.

\section{Experimental Setup}

\subsection{Datasets}
\label{sec:datasets}

We used three widely used benchmark datasets to evaluate our proposed approach: Berlin Kudamm~\cite{8202131,sunderhauf2015place}, Oxford Robot Car~\cite{maddern20171}, and Nordland~\cite{olid2018single}. All the datasets present challenging scenarios for VPR in terms of substantial viewpoint shift and drastic shift in visual appearance due to seasonal cycles or time of day, as described below:

\subsubsection{Berlin Kudamm} This dataset is downloaded from the crowd-sourced photo mapping platform Mapillary where two different perspectives of the same route are captured. In this dataset, confusing objects and dynamic distractors such as vehicles and pedestrians with homogeneous scenes lead to perceptual aliasing. The substantial viewpoint shift in particular adds to the complexity. The total traverse span is about $3$ Km, where the reference traverse contains $314$ frames and the query traverse has $280$ frames. In both the traverses, all the frames are geotagged.

\subsubsection{Oxford RobotCar} This dataset contains traverses of Oxford city captured during different seasonal cycles and times of the day. We use a subsampled version of the Overcast Summer and Autumn Night traverses\footnote{Originally 2015-03-17-11-08-44 and 2014-12-16-18-44-24 in~\cite{maddern20171}}. We use GPS data to subsample the original data to obtain a total traverse span of $1.5$ Km, resulting in a total of $213$ frames in summer and $251$ frames during the night, with frame spacing of approximately $5$-$6$ meters. The Overcast Summer and Autumn Night traverse provide a drastic shift in visual appearance due to season and time of day.

\subsubsection{Nordland} This dataset captures a $728$ km train journey during different seasonal cycles. We use the Summer and Winter traverse for our experiments. We use the first $7000$ images from both the traverses, which are uniformly subsampled to obtain $1000$ images for the reference summer traverse and $467$ for the query winter traverse. The combination of widespread vegetation and occasional unique objects in this dataset presents a challenging scenario on top of extreme appearance variations.

\subsection{Evaluation}
\label{sec:Metric}
We use Recall$@1$ as the performance metric since the output of a VPR system can be typically employed for precise 6-DoF SLAM/localization~\cite{zaffar2021vpr}. For a given localization radius, Recall$@K$ is defined as the ratio of correctly retrieved queries within the top K predictions to the total number of queries. We use a ground truth localization radius of $50$ meters, $45$ meters and $1$ frame respectively for Berlin, Oxford and Nordland datasets. For our place-specific utility estimation, we use top $10$ clusters and for the combined environment- and place-specific system, we use top $X'-1$ clusters, where $X'$ is the number of useful clusters determined by the environment-specific system. We provide full parameter sweeps in the results section for sensitivity analysis.

\begin{table*}
\centering
\caption{Quantitative Results: Performance comparison on three benchmark datasets.}
\scalebox{0.9}{
\begin{tabular}{llllllllll}
\toprule
         & \multicolumn{3}{c}{Berlin}                     & \multicolumn{3}{c}{Oxford}                     & \multicolumn{3}{c}{Nordland}                   \\
\cmidrule{2-10}
                        \textbf{Methods} & Recall$@1$       & Storage       & Time          & Recall$@1$       & Storage       & Time          & Recall$@1$       & Storage       & Time          \\
\cmidrule(lr{0.75em}){1-1} \cmidrule(lr{0.75em}){2-4} \cmidrule(lr{0.75em}){5-7} \cmidrule(lr{0.75em}){8-10} 

NetVLAD~\cite{arandjelovic2016netvlad}                 & 38.21          & -             & -             & 46.61          & -             & -             & 9.21           & -             & -             \\
Vanilla SuperPoint (SP)~\cite{detone2018superpoint} & 46.07          & 1             & 1             & 72.11          & 1             & 1             & 14.99          & 1             & 1             \\
Semantic Consistency    & 44.64          & 0.66          & 0.83          & 64.14          & 0.81          & 0.90          & 13.91          & \textbf{0.58} & \textbf{0.90} \\
Cluster Consistency     & 43.21          & 0.49          & 0.74          & 58.96          & 0.62          & 0.80          & 11.99          & 0.69          & 0.96          \\
\textit{Ours}: ES Utility         & \textbf{50.36} & 0.79          & 0.88          & \textbf{74.10} & 0.53          & 0.76          & \textbf{16.06} & 0.96          & 0.99          \\
\textit{Ours}: PS Utility       & 47.14          & \textbf{0.48} & \textbf{0.73} & 69.32          & \textbf{0.37} & \textbf{0.71} & 14.56          & 0.90          & 0.97          \\
\textit{Ours}: ES + PS Utility    & 49.64          & 0.70          & 0.84          & \textbf{74.10} & 0.48          & 0.74          & \textbf{16.06} & 0.94          & 0.98          \\

\midrule

Vanilla SP + SuperGlue (SG)~\cite{sarlin2020superglue} & 59.64          & 1             & 1             & \textbf{86.45}          & 1             & 1             & 20.34          & 1             & 1             \\
\textit{Ours}: ES Utility         & \textbf{61.07} & 0.79          & 0.88          & 86.06 & 0.53          & 0.76          & 20.12 & 0.96          & 0.99          \\
\textit{Ours}: PS Utility        & 51.43          & \textbf{0.48} & \textbf{0.73} & 82.86          & \textbf{0.37} & \textbf{0.71} & 20.34          & \textbf{0.90}          & \textbf{0.97}          \\
\textit{Ours}: ES + PS Utility    & 59.64          & 0.70          & 0.84          & 86.06 & 0.48          & 0.74          & \textbf{20.98} & 0.94         & 0.98          \\

\midrule

HF-Net's MobileNetVLAD~\cite{sarlin2019coarse}                 & 35.36          & -             & -             & 60.55          & -             & -             & 16.91           & -             & -             \\
Vanilla HF-Net  & 46.78          & 1             & 1             & \textbf{86.00}          & 1             & 1             & 27.83          & 1             & 1             \\
\textit{Ours}: ES Utility         & \textbf{48.57} & 0.79          & 0.86          & 84.46 & 0.52          & 0.77          & \textbf{29.34} & 0.97          & 0.99          \\
\textit{Ours}: PS Utility        & 45.35          & \textbf{0.63} & \textbf{0.79} & 81.67          & \textbf{0.47} & \textbf{0.75} & 26.34          & \textbf{0.90}          & \textbf{0.97}          \\
\textit{Ours}: ES + PS Utility    & 47.85          & 0.70          & 0.81          & 85.66 & \textbf{0.47}          & \textbf{0.75}          & 28.69 & 0.95          & 0.98          \\

\bottomrule
\end{tabular}
}
\label{tab:mainResult}
\end{table*}

\begin{figure*}
\centering
\begin{tabular}{cccc}
\includegraphics[width=0.17\textwidth]{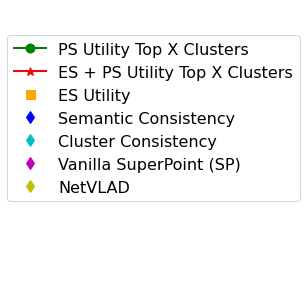}
&
\includegraphics[width=0.2\textwidth]{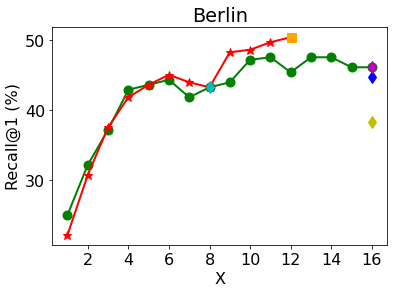}
& 
\includegraphics[width=0.2\textwidth]{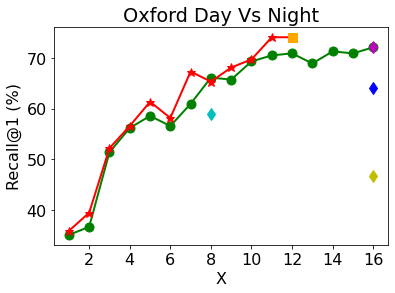}
&
\includegraphics[width=0.2\textwidth]{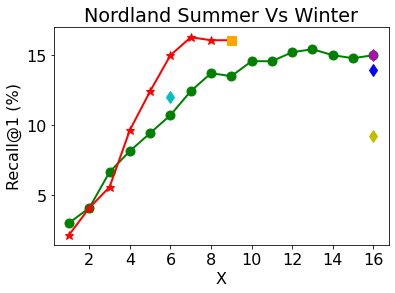} \\
\end{tabular}
\caption{Performance Vs Top-X cluster keypoints. Results displayed via single markers at $X=16$ use all the clusters and their keypoints (except Semantic Consistency).}
\label{fig:mainResult}
\end{figure*}

\subsection{Baseline Comparisons}
\label{sec:Details}
We use vanilla NetVLAD (Pitts30K trained~\cite{arandjelovic2016netvlad}), vanilla SuperPoint~\cite{detone2018superpoint} and vanilla SuperPoint + SuperGlue as baseline in the results, represented as \textit{NetVLAD}, \textit{Vanilla SuperPoint (SP)}, and \textit{Vanilla SP + SuperGlue (SG)} respectively. In all our local feature based methods including other baselines (described below), we use NetVLAD top-20 candidates to select the final match using local feature matching.

We also provide two additional baselines: Semantic Segmentation Consistency and Cluster Consistency, which employ human-level semantics in place of our ES \& PS utility in the proposed hierarchical VPR framework. These baselines are based on previous work leveraging human-level semantics~\cite{naseer2017semantics,garg2018lost,gawel2018x} or cluster-based fine-grained semantics~\cite{larsson2019fine}, where semantic or cluster label consistency across reference and query is used to improve the matching framework.

\begin{figure}
\centering
\begin{tabular}{cc}
\includegraphics[width=0.2\textwidth]{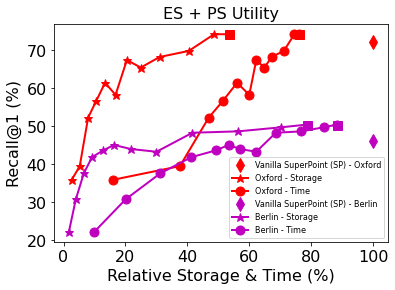}
&
\includegraphics[width=0.2\textwidth]{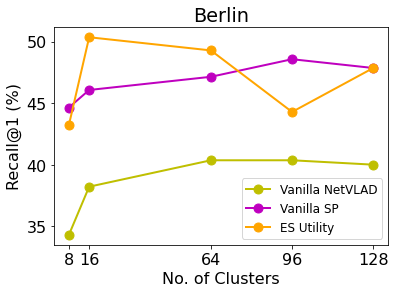}
\\
(a) & (b) \\
\end{tabular}
\caption{(a) Performance of ES + PS Utility with respect to storage and compute time relative to vanilla SuperPoint for Oxford and Berlin, and (b) performance variations on Berlin with respect to the vocabulary size, using Vanilla NetVLAD, Vanilla SuperPoint and ES Utility.}
\label{fig:StorageT_Vocab}
\end{figure}

For the Semantic Consistency baseline, we generate semantic segmentation masks for all reference and query images based on the Cityscapes~\cite{cordts2016cityscapes} scheme. We then filter the SP keypoints to only retain points belonging to buildings, vegetation, and roads for Oxford and Berlin (this particular choice of semantic classes is similar to the selection by~\cite{gawel2018x,garg2018lost}); and the ones belonging to buildings, vegetation and terrain for Nordland. In~\cite{garg2018lost}, implicit keypoint correspondences are first obtained and then semantic label consistency is imposed. For the Semantic Consistency baseline considered in this paper, we first obtain the keypoints from the chosen semantic classes and then use the vanilla SuperPoint's feature matching pipeline to find keypoint correspondences. Similarly, for the Cluster Consistency baseline, we select the clusters corresponding to the aforementioned semantic classes respectively for all three datasets based on visual inspection.

Finally, we also demonstrate the use of our proposed method on an existing hierarchical 6-DoF localization pipeline HF-Net~\cite{sarlin2019coarse} but in the context of VPR. In Table~\ref{tab:mainResult}, we include results for \textit{HF-Net's MobileNetVLAD} as a global descriptor, \textit{vanilla HF-Net} (using their global and local descriptors in accordance with the vanilla SP pipeline) and proposed utility-based keypoint filtering applied to vanilla HF-Net's local descriptors.

\begin{figure}
\centering
\begin{tabular}{cccc}
\includegraphics[width=0.4\textwidth]{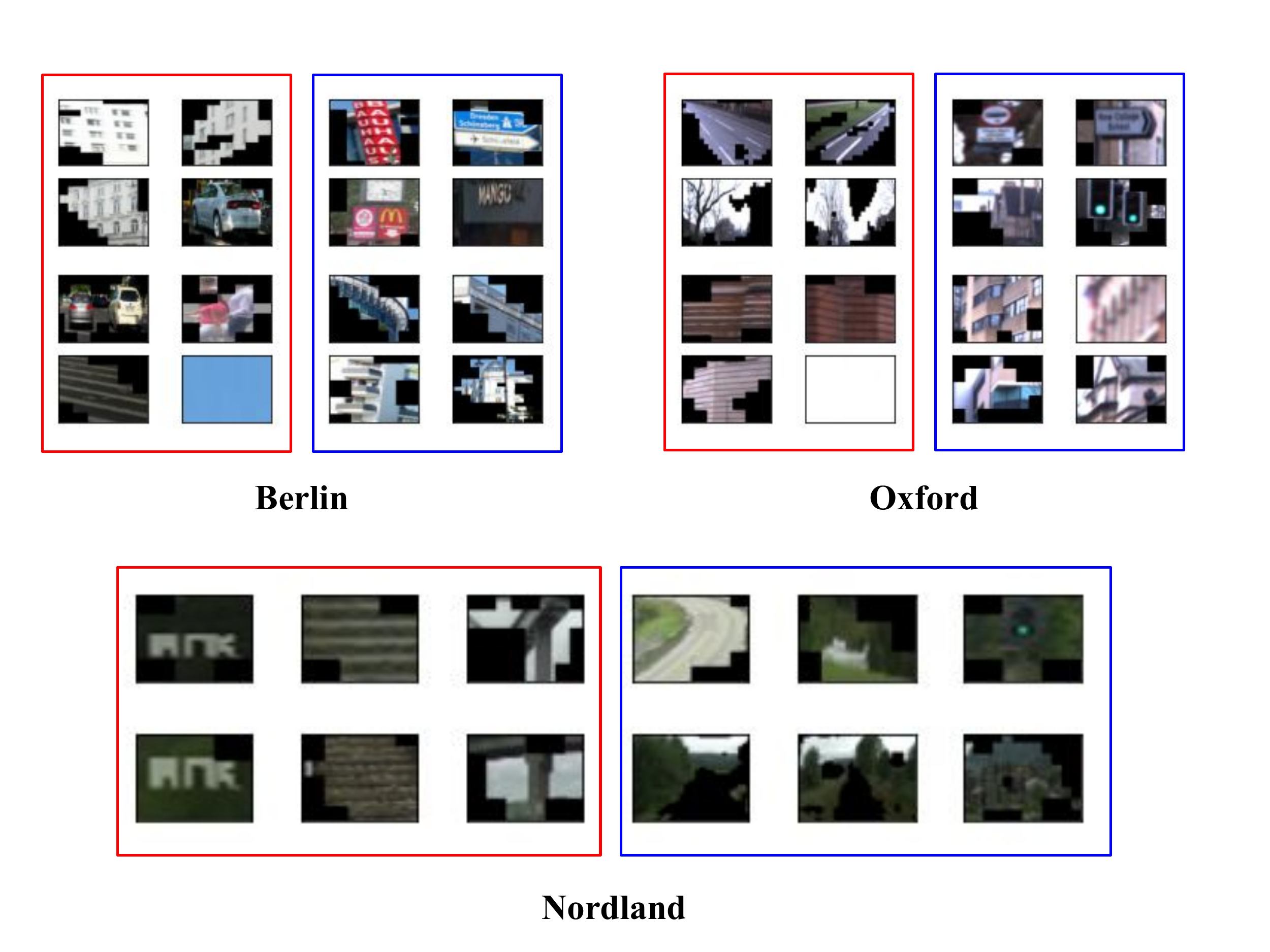} \\
\end{tabular}
\caption{Example patches from clusters with \textit{low} (red) \& \textit{high} (blue) environment-specific utility for Berlin, Oxford and Nordland.}
\label{fig:ES}
\end{figure}

\section{Results \& Discussion}

In this section, we first present the key quantitative results from testing the proposed framework on three benchmark datasets. We then provide a qualitative analysis with visualizations and insights from both Environment-Specific and Place-Specific cluster utility.

\subsection{Performance Characteristics}
\label{sec:Quant}

Table~\ref{tab:mainResult} and Fig~\ref{fig:mainResult} show the performance of the proposed pipeline on all three benchmark datasets. We also present all the seven baselines' performance.

Across all datasets and all baseline matching systems (SP, SP+SG, HF-Net), it can be observed that the Environment-Specific (ES) utility results in improved recall in most cases with noticeable reduction in storage and compute time. On the other hand, the Place-Specific (PS) system performs close to its respective vanilla method but significantly reduces storage and compute requirements. Hence, the combined ES + PS method balances the trade-off between recall and efficiency advantages, leading to improved efficiency than ES alone and consistently superior recall performance as compared to the vanilla methods (with only exception being the Oxford dataset when using SP+SG and HF-Net).

Fig~\ref{fig:mainResult} shows the full performance curves for the PS and ES+PS systems for SuperPoint-based matching. Performance saturates at a relatively small number of top-X clusters. When employing a combined filtering approach based on both ES and PS Utility, this peak performance is improved and achieved rather earlier. 

\newcommand{\scaleOne}{0.08\textwidth}
\newcommand{\scaleColorBar}{0.06}
\begin{figure}
\centering
\begin{tabular}{llllc}
\includegraphics[width=\scaleOne]{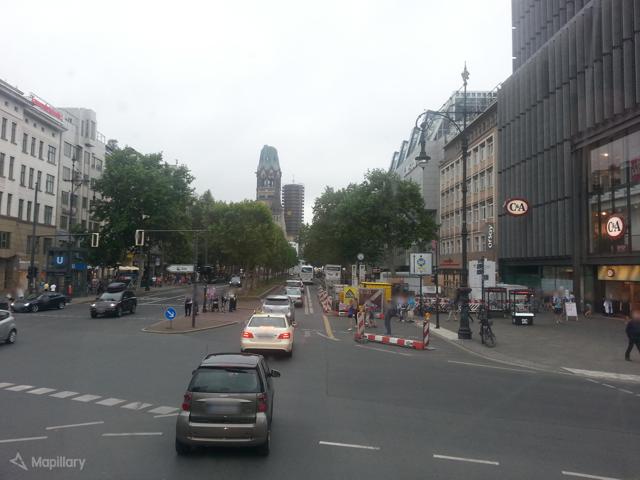}  
&
\includegraphics[width=\scaleOne]{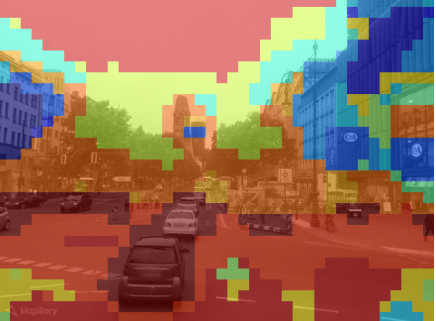}
&
\includegraphics[width=\scaleOne]{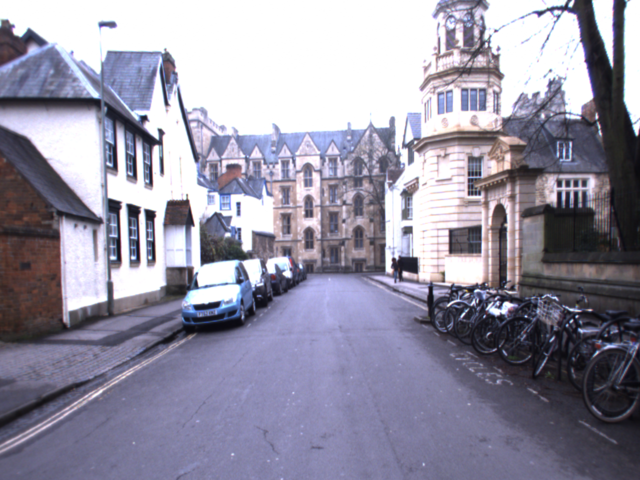}
&
\includegraphics[width=\scaleOne]{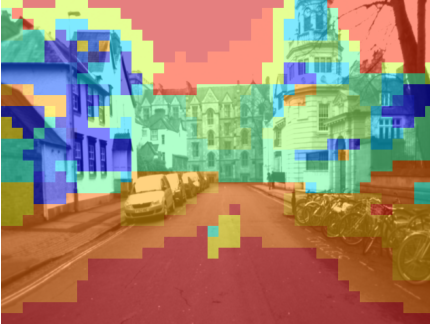}
&
\includegraphics[trim={0.1cm 0.1cm 0.1cm 0cm},clip,scale=\scaleColorBar]{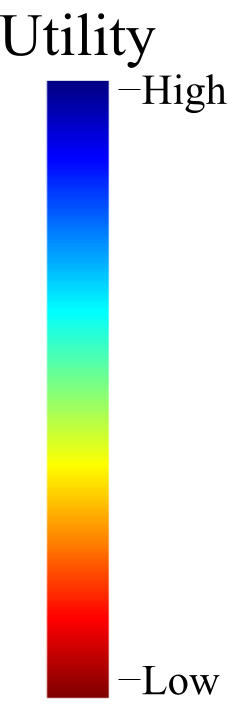} \\
\multicolumn{2}{c}{(a) Berlin} & \multicolumn{2}{c}{(b) Oxford} &                       \\
\includegraphics[width=\scaleOne]{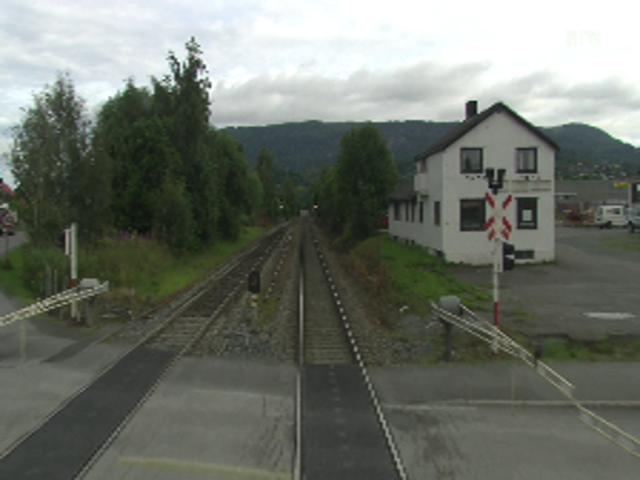}        & 
\includegraphics[width=\scaleOne]{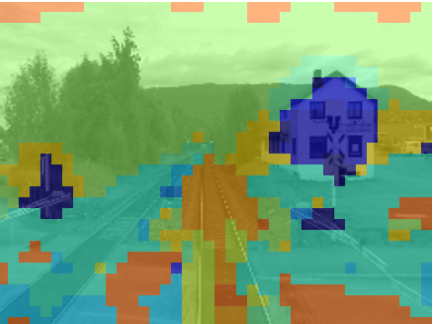}
& 
\includegraphics[width=\scaleOne]{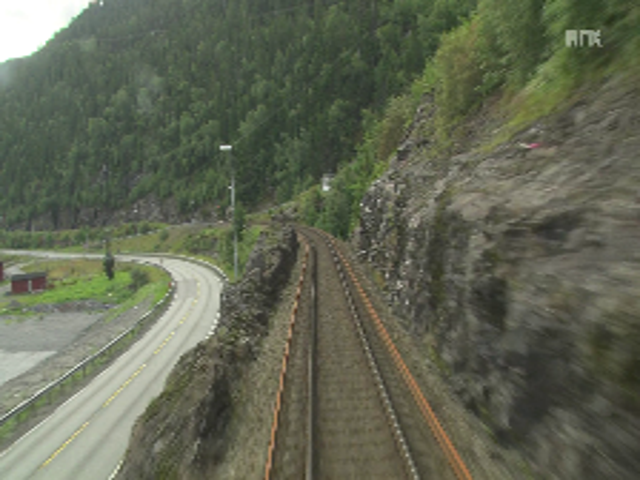}
& 
\includegraphics[width=\scaleOne]{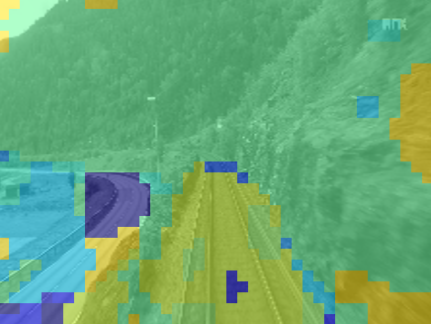}
&
\includegraphics[trim={0.1cm 0.1cm 0.1cm 0cm},clip,scale=\scaleColorBar]{results/colorbar.png} \\
\multicolumn{4}{c}{(c) Nordland Summer}                         &     
\end{tabular}
\caption{Visualizations of PS utility ranking of different clusters, where utility decreases from blue to green to red.}
\label{fig:PS}
\end{figure}

\subsubsection{Storage \& Compute Time Benefits}
\label{sec:StorageTime}

Fig \ref{fig:StorageT_Vocab}(a) shows the ES utility-based filtering resulting in a consistent reduction of the reference map size, measured in terms of total number of local descriptors stored. This leads to reduction in compute time. In particular, for the Oxford dataset, where perceptual aliasing is high due to day-night matching, the ES utility-based filtering leads to an increase in performance while only requiring storage of $55\%$ of the original descriptors.

Both the standalone PS and combined PS+ES system is able to retain near peak performance while reducing the number of Top-X clusters selected for PS utility. In particular for the Berlin dataset, PS+ES continues to outperform the vanilla system even with $40\%$ storage and $70\%$ compute-time requirements. At $20\%$ storage and $50\%$ compute-time, performance is still competitive with Vanilla SP. Similar trends can be observed for the Oxford and Nordland datasets. Further computational and storage gains are likely achievable with complementary methods including quantization, binarization, hashing and dimension reduction~\cite{arroyo2015towards,vysotska2017relocalization,jegou2010aggregating}. 

\subsubsection{Effect of Vocabulary/Cluster Size}
\label{sec:vocabSize}
To further understand the effect of the VLAD vocabulary size (number of clusters) on the proposed pipeline's performance, we present an ablation using $8$, $16$, $64$, $96$, and $128$ clusters in the proposed pipeline. Fig \ref{fig:StorageT_Vocab}(b) shows the performance of Vanilla NetVLAD, Vanilla SuperPoint and ES Utility methods on Berlin for varying numbers of clusters. It can be observed that the use of 16 clusters for ES offers relatively bigger jump in performance than other cluster size values, which even surpasses high baseline performance for cluster size 96 and 128. Although the distinctiveness of individual clusters increases with the vocabulary size, the results suggest that their relative utility ranking does not remain meaningful enough to noticeably improve performance.

\subsection{Qualitative Analysis of ES \& PS Utility}
\label{sec:Qual}
Our quantitative analyses presented above show how the proposed ES and PS utility methods achieve state-of-the-art performance while also reducing the overall storage and compute time requirements. In this section, we present insights and visualizations of segmentation masks obtained from ES, PS and their combined utility estimation. 

\subsubsection{Environment-Specific Utility}
\label{sec:SemES}

Prior work~\cite{naseer2017semantics,garg2018lost,gawel2018x} was based on an assumption that a handful of broad-level semantic classes such as buildings, vegetation, and roads are more important to VPR. Fig~\ref{fig:ES} depicts patches from low and high ES utility clusters for all three datasets. It can be observed that generic distractors such as cars, pedestrians and sky are discarded by ES utility, which helps in improving performance and is in line with a broad semantics-based utility. 
However, a cluster representing ``a large white planar patch with a window grid'' in the Berlin dataset (top left, red) and ``large road patches'' in the Oxford dataset (top right, red) are deemed to have low utility due to their frequent occurrence, leading to high perceptual aliasing. This demonstrates that determining feature utility for VPR based on a broad-level semantic class is not sufficient, thus fine-grained representations (often a subset of a broad-level semantic class) specific to an environment can effectively determine cluster utility to improve performance.

In the Nordland dataset, there is a small watermark at the top-right corner of the image, which often leads to false local keypoint matches for vanilla SuperPoint. While this specific example could be trivially filtered, it demonstrates a key property of the methods presented here. As shown in Fig~\ref{fig:ES}, in the Nordland cluster patches, this frequently-occurring watermark is assigned to a low environment-specific utility cluster which represents ``text based visual elements".

\begin{figure}
\centering
\begin{tabular}{llllc}
\includegraphics[width=\scaleOne]{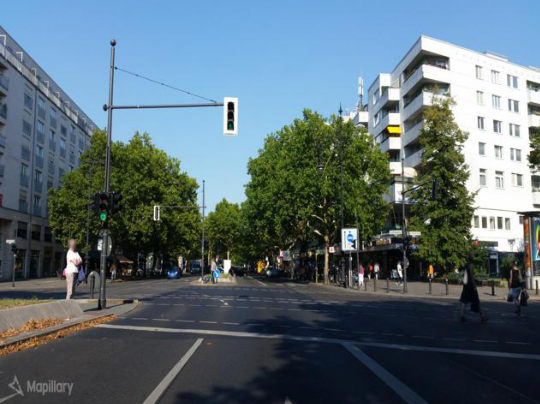}     &
\includegraphics[width=\scaleOne]{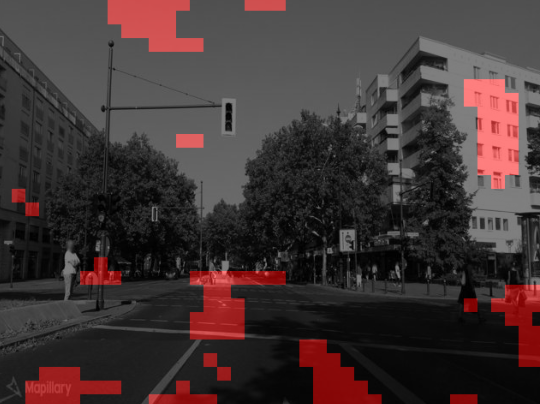} &
\includegraphics[width=\scaleOne]{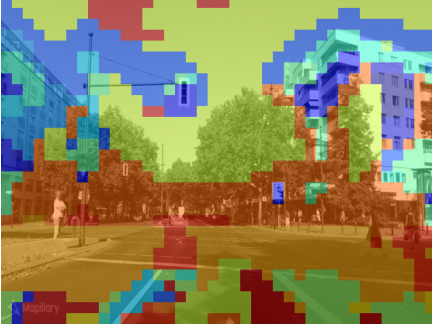}     &
\includegraphics[width=\scaleOne]{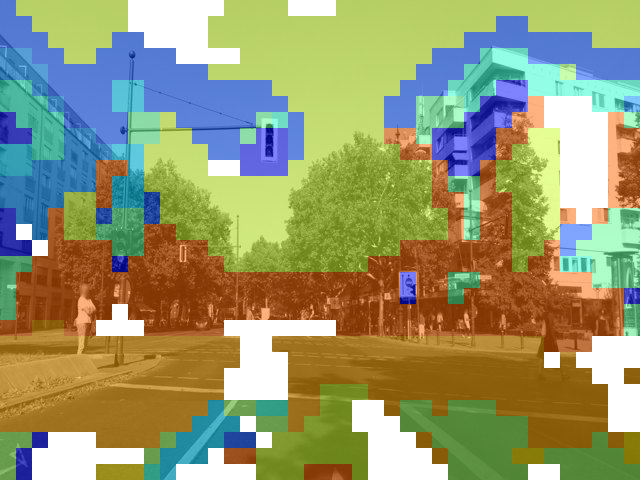}   &
\includegraphics[trim={0.1cm 0.1cm 0.1cm 0cm},clip,scale=\scaleColorBar]{results/colorbar.png} \\
\multicolumn{4}{c}{(a) Berlin}      & 
\\
\includegraphics[width=\scaleOne]{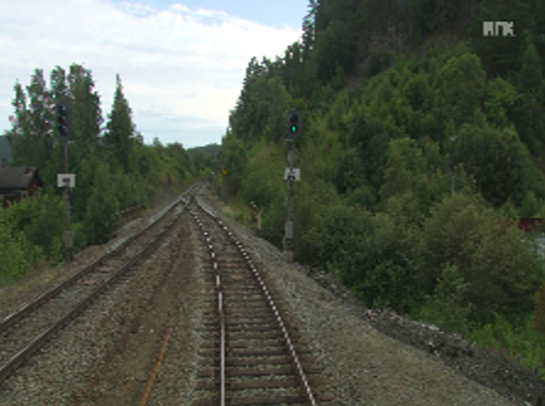}       &
\includegraphics[width=\scaleOne]{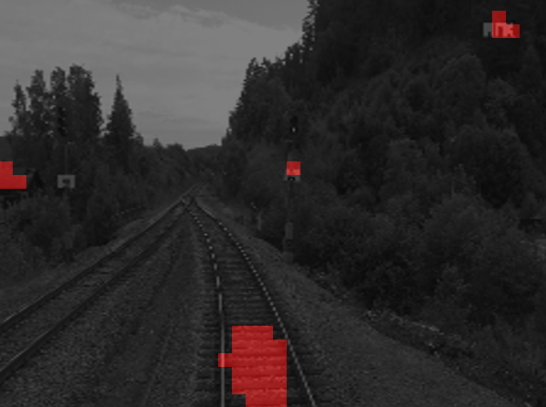}   &
\includegraphics[width=\scaleOne]{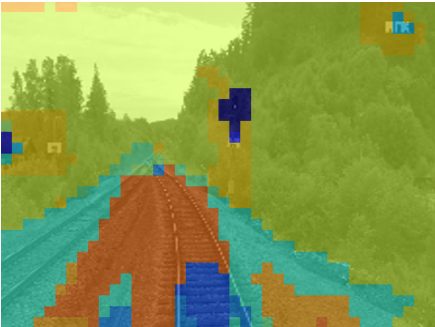}       & 
\includegraphics[width=\scaleOne]{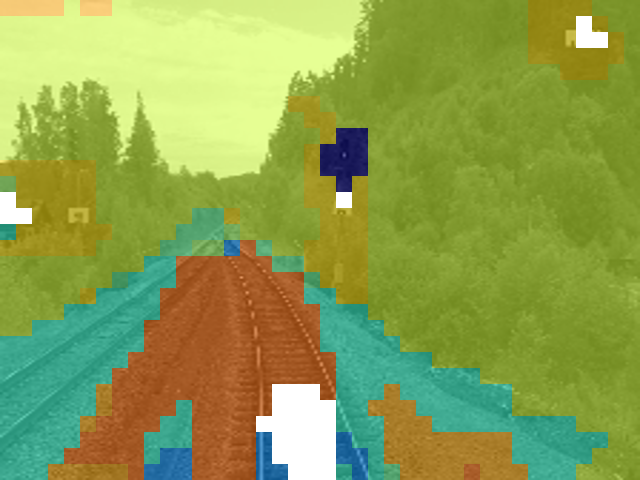}     &
\includegraphics[trim={0.1cm 0.1cm 0.1cm 0cm},clip,scale=\scaleColorBar]{results/colorbar.png} \\
\multicolumn{4}{c}{(b) Nordland Summer} &
\end{tabular}
\caption{Visualization of ES utility (second column), PS utility (third column) and ES+PS utility (last column). For ES visualizations, low utility is represented in red and high in gray..}
\label{fig:ES_PS}
\end{figure}

\subsubsection{Place-Specific Utility}
\label{sec:SemPS}
Fig \ref{fig:PS} shows the importance ranking visualizations based on PS utility, with blue being most useful and red the least. While the behavior of PS is similar to ES, notable examples can be observed in an open vegetative environment of Nordland, where an extra set of railway track (cyan colored in Fig~\ref{fig:PS}(c) left) and a part of road (blue and cyan colored in Fig~\ref{fig:PS}(c) right) specific to a particular place is marked as more important than other elements. Similarly, in Fig~\ref{fig:PS} c) right, the road visible on the left side is a highly-informative place-specific cue and is considered as high utility by the PS system. As a particular characteristic of the Nordland dataset, we observed that pixels belonging to vegetation, clouds and sky sometimes emerged as a single cluster and thus cumulatively assigned a middling place-specific utility. This could potentially be mitigated through a dynamic selection of cluster size per place and can be explored in future work.

\subsubsection{Combined ES + PS Utility}
\label{sec:UnifiedQual}

Fig \ref{fig:ES_PS} shows how ES and PS Utility complement each other and their combination leads to better utility ranking. For instance, in Berlin, both PS and ES utility consider buildings as important on both sides of the road but ES discards the window panes as discussed previously, leading to improved ES+PS performance. Similarly, as shown in Fig~\ref{fig:ES_PS} for Nordland, PS assigned high utility to a small part of the railway track and watermark but ES utility discarded them. Hence, in the combined ES+PS system, the PS utility-based filtering becomes more robust to perceptual aliasing while also avoiding transient errors.

\subsubsection{Case Study: Frequent Virtual Landmark}
\label{sec:CaseStudy}

In order to analyze the effect of uniqueness or frequency of an object in the reference traverse on utility estimation, we present a controlled experiment on the Oxford Summer day traverse. We introduce a unique virtual landmark (traffic sign) in the image such that it is well-aligned with the road across the traverse. We introduce this landmark in the traverse at $t{^th}$ image to create four duplicate reference traverses where $t$ varies as $50$, $10$, $4$, and $1$ for \textit{Sparse, Moderate, High, and Dense} setting respectively. As the frequency of the added landmark increased, the PS and ES utility values of that particular cluster decreased. In Fig~\ref{fig:CaseStudy}, it can be observed that for the \textit{Sparse} setting, the landmark is marked as a salient object since it only appears at very few places across the traverse. Conversely, for the \textit{Dense} scenario, it can be observed that the landmark is marked as the least salient by both PS and ES utility estimation. This study shows the effectiveness of the proposed unsupervised method where regardless of any prior knowledge of the virtual landmark or specific training, the utility is correctly estimated.

\section{Conclusion}

Discriminatively identifying individual places from a set of already-seen images is a critical and challenging problem for VPR. Estimating the uniqueness of visual cues and their relevance to VPR is crucial in the context of this problem. In this research, we proposed a novel approach to deduce the utility of visual cues `specific' to an environment and a particular place, unified through a pipeline that guides keypoint filtering at the local feature matching stage. Our proposed pipeline leads to consistent state-of-the-art performance on three standard benchmark datasets exhibiting challenging appearance change and viewpoint shift, while simultaneously reducing the storage and compute time requirements.

\newcommand{\scaleTwo}{0.08\textwidth}
\newcommand{\scaleColorBarTwo}{0.06}
\begin{figure}
\centering
\begin{tabular}{cc|ccc}
\includegraphics[width=\scaleTwo]{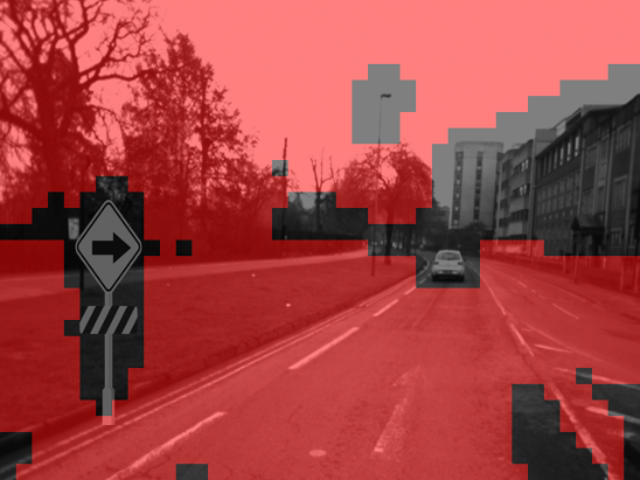}
&
\includegraphics[width=\scaleTwo]{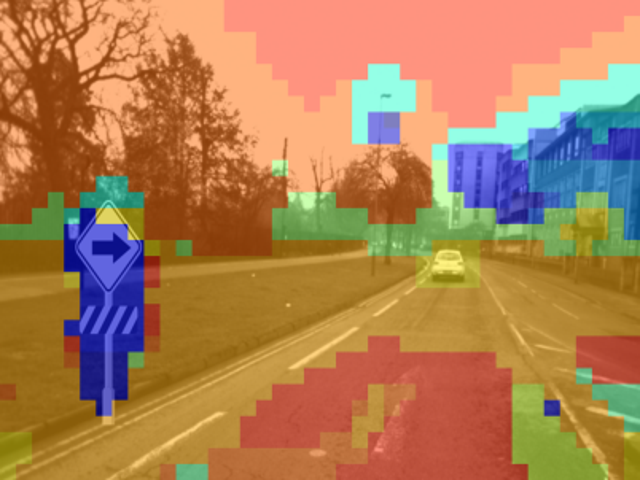}
&
\includegraphics[width=\scaleTwo]{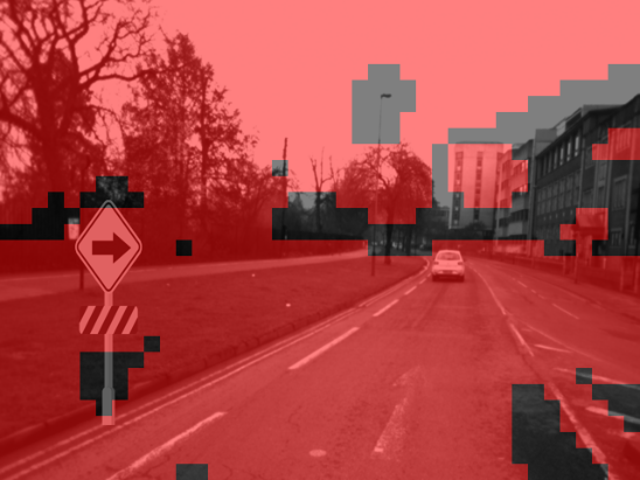}
&
\includegraphics[width=\scaleTwo]{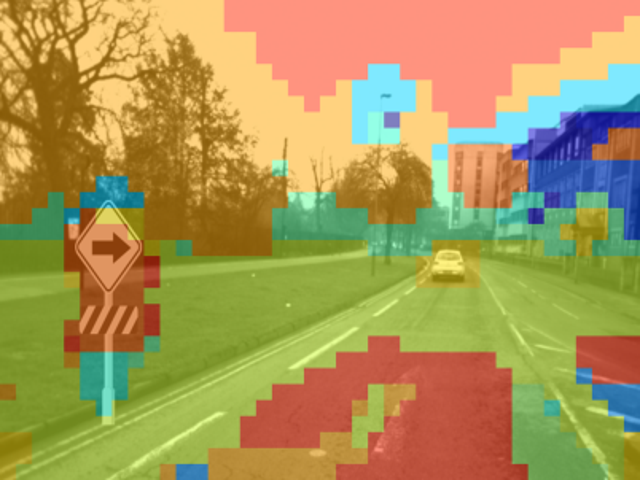} &
\includegraphics[trim={0.1cm 0.08cm 0.1cm 0cm},clip,scale=\scaleColorBarTwo]{results/colorbar.png}\\
\multicolumn{2}{c}{(a) Sparse} & 
\multicolumn{2}{c}{(b) Dense} \\
\end{tabular}
\caption{Visualization of ES (left) and PS (right) utility ranking when a virtual landmark is placed (a) sparsely or (b) densely in the traverse. For ES visualizations, low utility is represented in red and high in gray.}
\label{fig:CaseStudy}
\end{figure}

A number of areas are of interest for further investigation. We employed a contrastive learning approach here but there may be other better suited learning schemes. Moving beyond the two utility measures investigated here, it may be profitable to learn a much larger number of measures of utility, for example learning a measure of utility for local areas that lies partway between specific places and whole environment measures. The ultimate ideal number of utility measures may depend on their complementarity, which could also be assessed. While we have shown here that finer grained semantic segmentation below broad classes like road or building may have particular utility for VPR, further research could investigate the relationship between human-defined categories and those which are most useful for VPR. Collectively, future work in this area will help further improve the capabilities of these systems while also bridging the divide between human navigation and autonomous navigation systems, with potential additional benefits in areas like human robot interaction.


\maxdeadcycles=1000
\bibliographystyle{IEEEtran}
\bibliography{ref}

\end{document}